\documentclass[runningheads]{llncs}

\usepackage{eccv}

\usepackage{eccvabbrv}

\usepackage{graphicx}
\usepackage{booktabs}
\usepackage{threeparttable}
\usepackage{array} 
\usepackage{booktabs}   %
\usepackage{tabularx}   %
\usepackage{siunitx}    %

\usepackage[accsupp]{axessibility}  %

\usepackage{hyperref}

\usepackage{orcidlink}
\usepackage{xcolor}
\definecolor{darkgreen}{RGB}{0,100,0}
\definecolor{darkred}{RGB}{139,0,0}

\begin{document}

\title{MistyPilot: Enabling Social-Robot Control through Multi-Agent LLM Skill Orchestration}

\title{MistyPilot: Enabling Social-Robot Control through Multi-Agent LLM Skill Orchestration}
\titlerunning{MistyPilot}

\author{
Xiao Wang\textsuperscript{*}\,\orcidlink{0009-0000-0511-8495}
\and
Lu Dong\textsuperscript{*}\,\orcidlink{0009-0007-4036-7690}
\and
Ifeoma Nwogu\,\orcidlink{0000-0003-1414-6433}
\and
Srirangaraj Setlur\,\orcidlink{0000-0002-7118-9280}
\and
Venu Govindaraju\,\orcidlink{0000-0002-5318-7409}
}

\authorrunning{X. Wang et al.}

\institute{
State University of New York at Buffalo\\
\url{https://wangxiaoshawn.github.io/MistyPilot.html}
}

\maketitle

\begingroup
\renewcommand{\thefootnote}{\fnsymbol{footnote}}
\footnotetext[1]{Equal contribution.}
\endgroup

\begin{abstract}
Programming small social robots from natural-language instructions requires more than invoking isolated APIs. Interactive tasks combine reactive physical behaviors with stateful social behaviors, while existing interfaces often require developers to manually compose APIs into skills, configure their parameters, bind sensor events to skills, and manage task states at runtime. 
We present MistyPilot, a multi-agent LLM framework that interprets high-level natural-language instructions and orchestrates the corresponding skills on the Misty social robot. 
A Task Router dispatches each instruction to one of two specialized agents: a Physically Interactive Agent for sensor-triggered robot control and direct skill invocation, and a Social Interaction Agent for dialogue-oriented task-state management and context-dependent multimodal response generation. 
To improve efficiency, the Social Interaction Agent reuses previously generated results when applicable and invokes full generation otherwise. 
We evaluate MistyPilot on five component-level suites, with sensor bindings and skill invocations executed on the physical Misty robot, and a preliminary user study with 12 participants. MistyPilot attains high accuracy on routing, sensor-skill binding, task-state parsing, result reuse, and skill extension up to 100 skills, and lower variance than an otherwise identical single-agent baseline, while participants report positive perceptions of usability and interaction quality. The code will be made publicly available via the project page.

\end{abstract}

\section{Introduction}
\begin{figure}[t]
    \centering
    \includegraphics[width=1.0\linewidth]{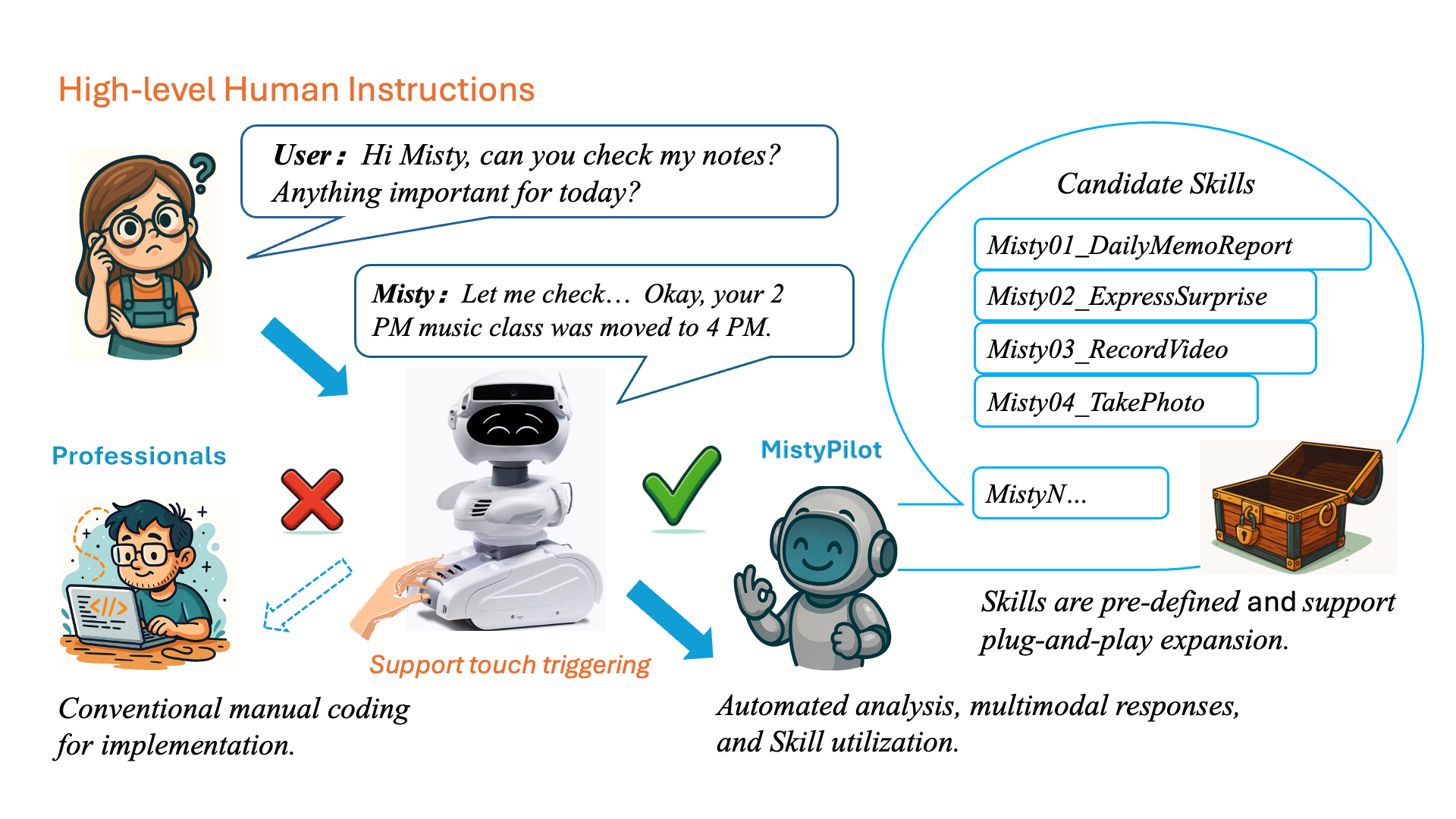}
    \caption{Overview of MistyPilot workflow for interpreting high-level human instructions. Instead of requiring professionals to hand-code robot behaviors and deploy features on the Misty robot, MistyPilot parses natural-language instructions, analyzes the task, selects and parameterizes skills from its library, and executes them on the Misty Robot.}
    \label{fig:intrograph}
\end{figure}

\label{sec:intro}
Social robots are increasingly used for everyday assistance, education, and social interaction~\cite{wang2025social,jackson2009teachers,gonzalez2025social,romero2024using,sawik2023robots,zhao2025social,ghafurian2025systematic,chen2020social}. 
A key challenge for social-robot platforms is enabling everyday users, rather than only programmers, to control robot behavior. This accessibility is particularly important in assistive settings, where users may have little or no technical background. Platforms such as Misty partially address this challenge by exposing open APIs for speech, vision, head and arm motion, and touch sensing~\cite{wang2025automisty,ciuffreda2025design}. However, using these capabilities still requires developers to implement robot behaviors as code by composing APIs into skills, configuring their parameters, binding sensor events to skills, and managing task states at runtime~\cite{eckert2025programming}.
Real user requests are difficult to enumerate in advance because they are often diverse, open-ended, and subject to change during interaction. 
In a single session, for instance, a user might first say, ``Greet me whenever I tap your head,'' and then ask the robot, ``Tell me a bedtime story,'' later requesting a different ending and asking to hear the story again.
The first is a reactive physical-interaction request that binds a sensor event to a robot skill, whereas the second is a stateful social-interaction request whose execution depends on dialogue history and task progress. 
Supporting both forms of interaction requires the robot to interpret user instructions and orchestrate the corresponding skills at runtime. Yet current workflows for social robots such as Misty still require developers to anticipate and manually implement these interaction patterns in advance, leaving non-expert users with limited ability to create or modify them at runtime.

These limitations arise from three technical gaps in current social-robot workflows. 
First, reactive physical interaction requires the robot to invoke skills directly and persist sensor-skill bindings so that they remain available across system restarts.
As the skill library grows, integrating new skills typically requires developers to modify the robot’s control logic manually. On Misty, AutoMisty~\cite{wang2025automisty} reduces the effort of creating individual skills by using an LLM to generate control code, but it does not discover, invoke, or orchestrate those skills at runtime. 
General LLM-based tool agents~\cite{schick2023toolformer,yao2023react,qin2023toolllm} provide flexible mechanisms for selecting and invoking tools from natural-language instructions at runtime. However, general-purpose tool orchestration remains largely underexplored on Misty social robots, where execution must additionally account for robot-specific skills and sensor-triggered interactions.
Second, stateful social interaction requires the robot to maintain task progress across dialogue turns. Users may revise an ongoing request, refer to earlier results, or ask the robot to repeat an updated task. Treating each utterance as an independent request makes it difficult to preserve dialogue context and execute evolving multi-step tasks consistently. Current workflows provide limited support for explicit dialogue-oriented task-state management.
Beyond state management, social responses should also be generated efficiently and expressed through coordinated robot modalities. Generating similar results from scratch for every request introduces redundant computation and latency. At the same time, basic text-to-speech and weakly coordinated motion can make robot responses feel mechanical. An effective system should therefore reuse suitable prior results when possible, invoke full generation when necessary, and produce context-dependent responses across speech, motion, and facial cues.

To address these limitations, we present MistyPilot, a multi-agent LLM framework that interprets high-level natural-language instructions and orchestrates the corresponding skills on the Misty robot at runtime. Its role-specialized architecture separates reactive physical interaction from stateful social interaction, enabling each type of task to be handled within a dedicated state and skill space. Our main contributions are summarized as follows:
\begin{itemize}

\item \textbf{A role-specialized architecture for reactive and stateful interaction.}
We introduce a multi-agent architecture that explicitly separates reactive physical interaction from stateful social interaction. A \textit{Task Router} directs each instruction to either a \textit{Physically Interactive Agent (PIA)} or a \textit{Social Interaction Agent (SIA)}, avoiding the need to combine routing, sensor-triggered control, and dialogue management within a single agent context.

\item \textbf{Specialized runtime orchestration for physical and social behaviors.}
The \textit{PIA} supports direct skill invocation, persistent sensor-skill bindings, and runtime integration of new skills. The \textit{SIA} maintains explicit task states across multi-turn dialogue, reuses previously generated results when applicable, falls back to full generation otherwise, and produces context-dependent multimodal responses across speech, motion, and facial cues.

\item \textbf{Component-level evaluation and preliminary user feedback.}
We evaluate MistyPilot on the Misty robot using five component-level suites covering routing, sensor-skill binding, task-state parsing, result reuse, and skill extensibility, together with a preliminary in-person user study. Under controlled Misty-specific settings, the system demonstrates reliable component-level performance and positive participant perceptions of usability and interaction quality.

\end{itemize}

\section{Related Work}

\subsection{Tool-augmented and embodied LLM agents}
LLM agents have progressed from pure reasoning to tool use. Toolformer~\cite{schick2023toolformer} learns to call APIs within text generation, ReAct~\cite{yao2023react} interleaves reasoning and acting to query external tools, ToolLLM~\cite{qin2023toolllm} orchestrates many tools through tree-structured search, and Tulip Agent~\cite{ocker2024tulip} scales to large tool libraries via vector-store retrieval; benchmarks such as API-Bank~\cite{li2023api}, APIBench~\cite{patil2024gorilla}, and ToolAlpaca~\cite{tang2023toolalpaca} standardize their evaluation. In the physical world, AutoRT~\cite{ahn2024autort}, Odyssey~\cite{liu2024odyssey} pair tool use with perception and planning. However, these works are not designed for social robots; they mainly target software tool calling or robotic motion planning, leaving their application to social-robot interaction largely unexplored.

\subsection{Social Robots for Human–Robot Interaction}
Prior work on social robots has explored affect recognition in assistive and mental-health settings~\cite{kling2025social}, but generating emotionally expressive multimodal responses remains underexplored. A related challenge is enabling non-programmers to create and control robot behaviors. AutoMisty~\cite{wang2025automisty}, the closest work to ours, uses a multi-agent LLM system to generate executable Misty skills for non-programmers, but it does not invoke or orchestrate those skills at runtime. MistyPilot complements AutoMisty by using a given skill library to interpret natural-language instructions and select, bind, and invoke skills on the physical robot, while also managing dialogue task state and generating emotion-conditioned multimodal responses.

\section{PROBLEM FORMULATION}

To handle the open-ended interaction scenarios faced by social robots, we first formalize a Task Space $\mathcal{T}$. Given a natural language task instruction $T_i \in \mathcal{T}$ from this space, the objective is to construct an end-to-end tool orchestration pipeline that executes the specified task. To handle heterogeneous tasks, MistyPilot first employs a Task Router $R$ to dispatch the initial task $T_i$ to the appropriate tool agent $K^*$:
\begin{equation}
K^* = R(T_i), \quad K^* \in \{\textit{PIA, SIA}\}
\end{equation}

Once dispatched to the selected agent branch, the corresponding agent (SIA or PIA) parses the task $T_i$, automatically selects the required functions $f_j$ from a predefined tool library $\mathcal{F}$, and infers their corresponding parameter configurations $\theta_j \in \Theta_{f_j}$. Since completing a task typically requires invoking a series of functions, the final execution of task $T_i$ is represented as an ordered sequence of parameterized function calls:
\begin{equation}
\textit{Agent} (T_i) = \bigl( f_1(\theta_1), f_2(\theta_2), \dots, f_N(\theta_N) \bigr)
\end{equation}

Based on this formulation, the core objective of MistyPilot is to infer a valid sequence of function calls that satisfies the task requirements:
\begin{equation}
\textit{MistyPilot}(T_i) = \textit{Agent}_{K^*}(T_i) = \bigl( f_j(\theta_j) \bigr)_{j=1}^N
\end{equation}

\section{METHODOLOGY}

\begin{figure*}[t]
    \centering
    \includegraphics[width=\textwidth]{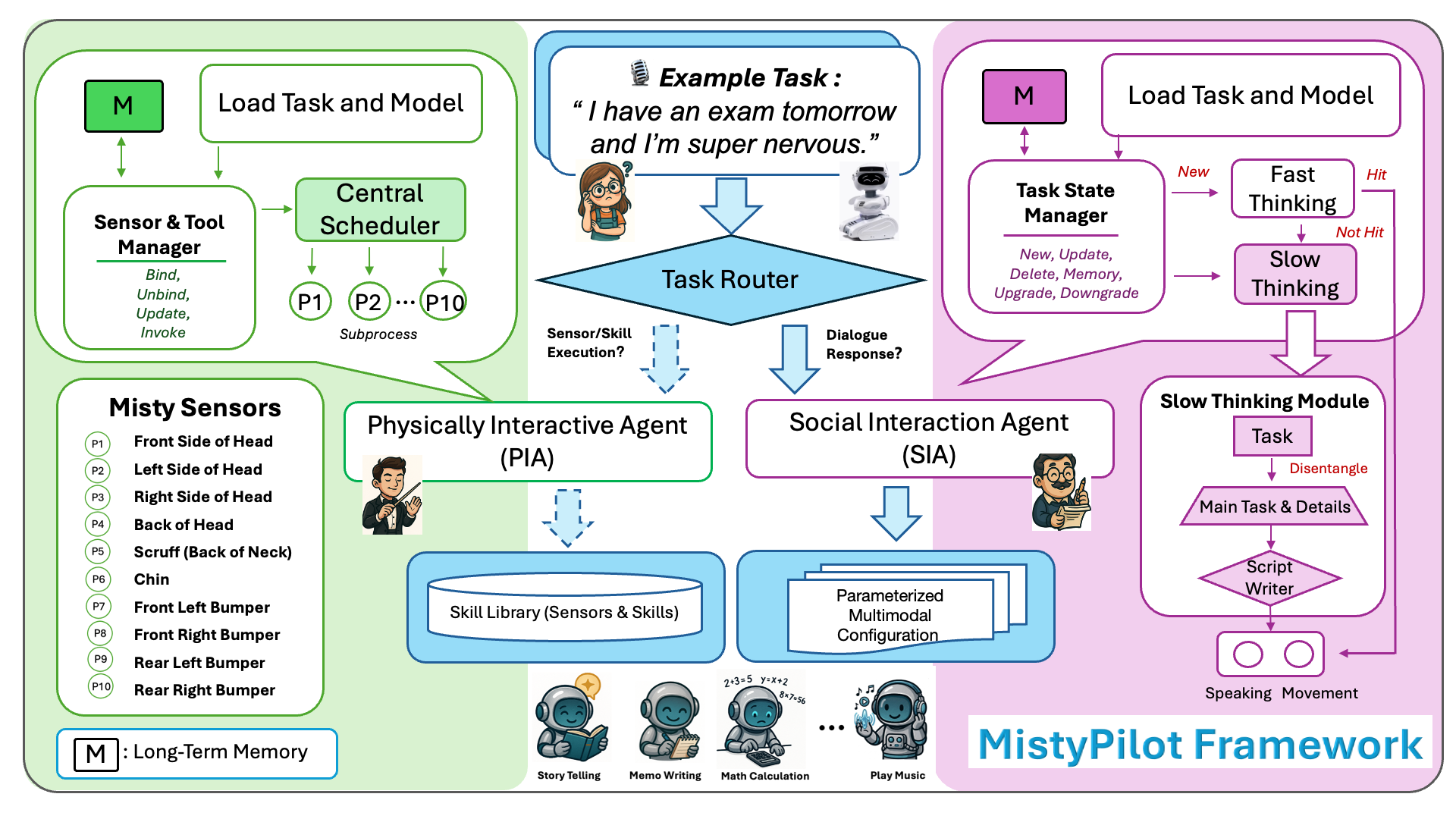}
    \caption{Overview of the MistyPilot framework. A \textit{Task Router} dispatches each instruction to the \textit{Physically Interactive Agent (PIA)} for sensor-skill binding and skill invocation, or to the \textit{Social Interaction Agent (SIA)} for dialogue. The \textit{SIA} tracks task state and responds through a fast path that reuses stored results, or a slow path that generates emotion-conditioned multimodal output.}
    \label{fig:overview}
\end{figure*}

To handle open ended human robot interactions, MistyPilot adopts a multi agent framework built on the tool agent paradigm~\cite{wang2024tools,qu2025tool,li2025review}, separating task routing, physical interaction, and social dialogue. A tool agent translates high level natural language instructions into parameterized skill calls and invokes registered tools to execute complex tasks, as illustrated in Fig.~\ref{fig:overview}. Crucially, MistyPilot adopts role separation at the architecture level rather than relying only on prompt based specialization. Unlike single agent systems that collapse the full context and all available skills into one agent, each agent in MistyPilot maintains its own localized state and skill space. The following subsections describe the design of the \textit{Task Router}, \textit{SIA}, and \textit{PIA}.

\subsection{Task Router}
Given a task, MistyPilot first employs the \textit{Task Router}, an LLM agent that acts as the central dispatcher. It analyzes the high-level natural-language instruction to determine the user's core intent. If the task involves sensor-triggered events or direct skill invocation, the router dispatches it to the \textit{PIA}; if the task is dialogue-oriented, it routes it to the \textit{SIA} to produce an emotion-conditioned multimodal response.

\subsection{Social Interaction Agent (SIA)}

When a dialogue-oriented task is dispatched to the \textit{SIA}, it is first processed by the \textit{Load Task and Model} module for state parsing and selection of the appropriate LLM version for reasoning. The \textit{Task State Manager (TSM)}, \textit{Fast Thinking}, and \textit{Slow Thinking} modules then coordinate to orchestrate the skills needed to complete the task.

\textbf{\textit{Load Task and Model}:}
This module converts user instructions into a task-state representation via an LLM and a schema-constrained system prompt. It maps natural-language intent to a predefined action set: \texttt{NEW}$\langle \mathit{main\_task}, \mathit{details}\rangle$, \texttt{UPDATE}$\langle \mathit{details}\rangle$, and \texttt{DELETE}$\langle \mathit{details}\rangle$ support task-state tracking and modification; \texttt{UPGRADE} and \texttt{DOWNGRADE} switch between a higher-capacity and a lightweight LLM for subsequent script generation; and \texttt{MEMORY} supports long-term information persistence upon user request.

\textbf{\textit{Task State Manager (TSM)}:}
Prior studies show that LLMs accumulate faithfulness errors during iterative dialogue summarization, particularly in long-context settings~\cite{liu2024lost,peysakhovich2023attention,maynez2020faithfulness,chang2023booookscore}.To address this and stabilize behavior, we use an explicit, editable external task-state memory to store and manage task states, reducing dependence on full-context recomputation.  Each update is atomic and localized, leaving other parts unchanged. Table~\ref{tab:tsm_example} shows an example user prompt and its TSM state: the details are stored as atomic units ($d_1,..., d_n$), so an \texttt{UPDATE} that changes the return time to 6:00 PM modifies only $d_2$ while the rest remain unchanged.

\begin{table}[h]
\centering
\renewcommand{\arraystretch}{1.3}
\caption{An example task status managed by the TSM. An \texttt{UPDATE} to one detail modifies only the affected atomic unit ($d_2$); the other units stay unchanged.}
\label{tab:tsm_example}
\begin{tabularx}{\columnwidth}{@{}l !{\color{gray}\vrule} X !{\color{gray}\vrule} X@{}}
\toprule
\multicolumn{3}{@{}p{\columnwidth}@{}}{\textbf{User Input 1:} \textcolor{blue}{``Hi Misty, I'd like to plan a day trip to New York City for tomorrow. Please create an itinerary that allows me to enjoy the city and return to my hotel by 7:00\,PM.''}}\\
\multicolumn{3}{@{}p{\columnwidth}@{}}{\textbf{User Input 2:} \textcolor{red}{``No, no. I need to return by 6:00\,PM.''}}\\
\midrule
 & \textbf{TSM State} & \textbf{After }\texttt{UPDATE}\textbf{(return by 6\,PM)}\\
\midrule
\textit{main\_task} & Plan a day trip to New York City & \textcolor{gray}{(unchanged)}\\
\midrule
\textit{details} & $d_1$: date = tomorrow & \textcolor{gray}{(unchanged)}\\
 & $d_2$: return by 7:00\,PM & \textcolor{red}{\textbf{$d_2$: return by 6:00\,PM}}\\
 & $d_3$: goal = enjoy the city & \textcolor{gray}{(unchanged)}\\
\bottomrule
\end{tabularx}
\end{table}

\textbf{\textit{Fast and Slow Thinking}:}
We adopt a dual-channel strategy inspired by fast and slow cognition in human problem solving~\cite{kahneman2011thinking,pan2025survey,evans2008dual}, with \textit{Fast Thinking} (retrieve and reuse) as the default and \textit{Slow Thinking} (generate from scratch) as a fallback. The reusable memory $\mathcal{M}$ is populated on demand: when the user marks a result worth keeping, the \texttt{MEMORY} action stores it, indexed by its $\mathit{main\_task}$ embedding. In \textit{Fast Thinking}, we encode the current $\mathit{main\_task}$ into a query embedding $\mathbf{q}$ and compare it against the stored embeddings $\mathbf{x}$ in $\mathcal{M}$ by cosine distance, as defined in Eq.~\eqref{eq:threshold}. The embedding space can be chosen flexibly; we compare three different embedding spaces, which yield consistent retrieval (Table~\ref{tab:test4}). We set the threshold $\tau=0.4$ based on the observed distance distribution. If the minimum distance exceeds $\tau$, including when no relevant item has been saved, we treat the task as a retrieval miss and fall back to \textit{Slow Thinking} for reasoning and generation.

\begin{equation}
    \min_{\mathbf{x} \in \mathcal{M}} d_{cos}(\mathbf{q}, \mathbf{x}) \le \tau
    \label{eq:threshold}
\end{equation}

In \textit{Slow Thinking}, the \textit{Script Writer} orchestrates the task with utterance-level emotion conditioning. Each utterance $u_i$ is assigned a multimodal emotion label from eight categories based on the extended Ekman emotion taxonomy~\cite{dalgleish2000handbook}: \textit{Happiness}, \textit{Sadness}, \textit{Anger}, \textit{Fear}, \textit{Disgust}, \textit{Surprise}, \textit{Contempt}, and \textit{Neutral}. Table~\ref{tab:script_segments} illustrates this process.

We express emotions consistently across three modalities: text, motion, and voice. Text is generated via context-aware reasoning. For motion, Misty executes expressive behaviors (e.g., arm waving, head tilting, body swaying, and light flashing) using templates aligned with Ekman's basic emotions~\cite{dalgleish2000handbook}, with LED colors and facial expressions reinforcing discriminability and cross-modal consistency. For voice, we bypass the built-in TTS and use OpenAI-TTS for controllable speech synthesis (timbre, speaking rate, and intonation). It (i) adapts prosody and intonation to text semantics and punctuation, (ii) injects emotion-conditioned style by encoding affective attributes such as arousal, intensity, and prosodic patterns~\cite{goudbeek2010beyond}, and (iii) controls speaking rate conditioned on emotion labels. Following~\cite{goudbeek2010beyond}, we discretize speaking rate by arousal: $1.00$ (baseline), $0.95$ (low arousal), and $1.05$ (high arousal), yielding synchronized action--speech expressions. Finally, the \textit{SIA} delivers motion and voice to the \textit{Speaking} and \textit{Movement} modules for simultaneous execution on Misty.

\begin{table}[h]
\centering
\renewcommand{\arraystretch}{1.2}
\caption{Emotional attunement example by the \textit{Script Writer}}
\label{tab:script_segments}
\begin{tabularx}{\columnwidth}{@{}c X l@{}}
\toprule
\multicolumn{3}{@{}c@{}}{\textbf{Prompt:} \textcolor{blue}{``Tell me the story of the Three Little Pigs''}} \\
\midrule
\textbf{\#} & \textbf{Text} & \textbf{Emotion} \\
\midrule
1 & Three little pigs left home to build their own houses. & Neutral \\
2 & The first pig quickly built a house of straw. & Contempt \\
3 & The second pig put up a house of sticks with little effort. & Contempt \\
4 & The third pig worked diligently to lay strong bricks for a sturdy home. & Happiness \\
\midrule
\multicolumn{3}{@{}c@{}}{\dots} \\
\bottomrule
\end{tabularx}
\end{table}

\subsection{Physically Interactive Agent (PIA)}
Tasks involving sensor-triggered events or direct skill invocation are routed to the \textit{PIA}, which performs reactive, event-driven robot control by mapping sensor events to skills and executing them when triggered. The \textit{PIA} uses the \textit{Load Task and Model} module to interpret user intent and either (i) directly execute a skill or (ii) identify the target sensor and retrieve the corresponding skill from the \textit{Skill Library}. It then uses the \textit{Sensor \& Tool Manager} module to complete the sensor bindings and execute the skill, enabling seamless human--Misty interaction.

\textbf{\textit{Load Task and Model}:}
As in the SIA, this module uses an LLM with a schema-constrained prompt to convert user instructions into sensor- or skill-related commands, mapping natural-language intent to a predefined command set: \texttt{BIND}$\langle s,k\rangle$ binds skill $k$ to sensor $s$; \texttt{UPDATE}$\langle s,k\rangle$ rebinds sensor $s$ to a new skill $k$; \texttt{UNBIND}$\langle s\rangle$ clears all skills bound to sensor $s$; and \texttt{INVOKE}$\langle k\rangle$ runs skill $k$ once without binding. Here $s \in S$ is one of Misty's ten sensors (Fig.~\ref{fig:overview}, \textit{Misty Sensors}), and $k \in K$ is a callable skill (e.g., ``play relaxing piano music''). The skill set $K$ is loaded at runtime for plug-and-play expansion: at startup, the \textit{PIA} scans the local skill directory, parses each docstring into an inventory of skill names and capabilities, and injects this inventory into its system.

\textbf{\textit{Sensor \& Tool Manager (STM)}:}
\textit{STM}, through a \textit{Central Scheduler} module, maintains an independent worker process for each valid sensor and records their status in a process table (\textit{sensor}, \textit{PID}, \textit{bound skill}, \textit{Status}). The \textit{Central Scheduler} performs periodic active checks, and any inactive process is immediately restarted, keeping sensor bindings available. Table~\ref{tab:process_table} presents example entries of the \textit{STM} process table. The fields \textit{sensor}, \textit{PID}, and \textit{bound skill} denote the sensor name, process identifier, and the associated skill, respectively. The \textit{Status} field indicates whether a sensor's PID exists in the system process tree (\textit{Active}) or has terminated unexpectedly (\textit{Inactive}).

Upon MistyPilot startup, \textit{STM} scans the persisted process table, recovers the sensor-skill bindings, and automatically remounts them on Misty, providing persistent state and rapid recovery across sessions.

\begin{table}[h]
\centering
\renewcommand{\arraystretch}{1.2}
\caption{Example of STM process table entries}
\label{tab:process_table}
\begin{tabular*}{\columnwidth}{@{\extracolsep{\fill}}llll@{}}
\toprule
\textbf{Sensor} & \textbf{PID} & \textbf{Bound Skill} & \textbf{Status} \\
\midrule
Head Touch        & 1023 & Greeting        & \textcolor{darkgreen}{Active} \\
Scruff Touch      & 1098 & DailyMemoReport & \textcolor{darkgreen}{Active} \\
Front Left Bumper & 1120 & NodHead         & \textcolor{darkred}{Inactive} \\
\bottomrule
\end{tabular*}
\end{table}

\section{Component-Level Evaluation Suites}

To evaluate the MistyPilot framework, we curate five evaluation suites that target task routing, sensor binding, task-state parsing, retrieval reuse, and skill extensibility. Their construction follows a hybrid pipeline~\cite{wang2023self} that synthesizes realistic instructions while mitigating single-model bias: (1) we collect empirically grounded seeds from experts with substantial on-site Misty demonstration experience; (2) we expand these seeds using a multi-LLM ensemble (Gemini 2.5 Pro and GPT-5) under explicitly defined Easy/Hard criteria, with single-feature iterative prompting to control complexity; (3) we enforce diversity via ROUGE-L filtering, removing samples with overlap greater than 0.7; (4) we randomly sample from the instruction pool to form each suite; and (5) human annotators assign the ground-truth routing decisions, target skills, and parameter specifications. The LLM ensemble is used only to generate candidate instructions; all ground-truth labels are assigned by human annotators, enabling quantitative evaluation of MistyPilot's decision accuracy.

\textbf{Route100}: This suite comprises 100 task instructions designed to evaluate the routing capability of MistyPilot, specifically whether a task should be dispatched to the \textit{SIA} for a dialogue-oriented task or to the \textit{PIA} for sensor-triggered events or direct skill invocation. It is relatively balanced across both categories, with 58 \textit{SIA} tasks and 42 \textit{PIA} tasks. Each subset is further divided into \emph{Easy} cases (direct and straightforward instructions) and \emph{Hard} cases (implicit or composite conditions), enabling a fine-grained evaluation of routing accuracy under diverse natural-language inputs.

\textbf{SensorBind40}: This suite contains 40 single-turn instances spanning two categories: (i) sensor-binding commands and (ii) immediate single-skill invocation commands without sensor binding, curated to evaluate the \textit{PIA}'s routing accuracy and sensor-binding capability. The instances are split by difficulty: 20 \emph{Easy} cases that require either a single, immediate sensor-grounded response or the immediate invocation of a single skill (e.g., ``touch your head and make a cute sound'', or ``I'm doing home exercise, play some good workout music for me''), and 20 \emph{Hard} cases that require parsing and registering multiple concurrent sensor-trigger bindings within a single pass (e.g., ``when I tap your chin, take a photo; press your forehead to say hi; touch your right side to show sadness''). These multi-binding instructions are challenging because all bindings must be correctly parsed, registered, and executed without omission.

\textbf{TaskParser256}: This suite contains 40 multi-turn dialogues (256 turns total) spanning multiple domains, including daily-life assistance, planning, storytelling, and emotion-supportive interaction. It evaluates the \textit{SIA}'s proficiency in parsing user intent, dynamically managing task states, and executing system-level controls (e.g., switching to a more powerful model) based on explicit user feedback. The suite is split into 7 \emph{Easy} dialogues (28 turns) and 33 \emph{Hard} dialogues (228 turns). \emph{Easy} dialogues contain clear instructions with no more than four turns, while \emph{Hard} dialogues involve longer interactions with ambiguous references (coreference), topic shifts, and interleaved or evolving instructions, posing challenges for long-horizon reasoning and context-aware adaptation.
 
\textbf{FastThinking230}: This suite targets MistyPilot's \emph{fast-thinking} module and evaluates its retrieval accuracy under paraphrasing. It contains 230 commands derived from 46 canonical tasks, each expanded into five variants that preserve the same core task while varying surface details. The goal is to measure whether the system can correctly retrieve and reuse previously executed implementations under surface-level linguistic variations, thereby reducing latency and computational cost.

\textbf{SkillExtension100}: This suite evaluates how well MistyPilot scales to newly introduced skills, assessing \textit{PIA} skill extensibility under dynamically injected skills. New skills are introduced at four scales (30, 50, 70, and 100 skills), and each skill is provided solely by its docstring; the system is expected to select the appropriate skill based only on these docstring descriptions.

\section{Experiments}
To evaluate MistyPilot, we conduct experiments on \textit{Task Routing} correctness, \textit{PIA} performance, \textit{SIA} performance, fast-thinking retrieval, and skill extensibility using the corresponding evaluation suites. To account for the stochastic nature of LLM outputs, each configuration is run five times with the temperature fixed at 0.3, and results are reported as mean $\pm$ standard deviation. Since MistyPilot’s contribution lies in architectural role separation rather than prompt engineering alone, we compare it with a Single-Agent baseline that collapses the \textit{Task Router}, \textit{PIA}, \textit{SIA}, and all skills into one LLM agent. For a fair comparison, the baseline uses the same GPT-5-mini backbone, skill set, inputs, and evaluation metrics as MistyPilot. Thus, the only difference is the system architecture: specialized agents with isolated contexts and restricted skill spaces versus a single agent with all context and skills collapsed together.

\subsection{Evaluation of Task Routing Correctness}
We evaluate MistyPilot's \textit{Task Router} on the MistyPilot-Route100 suite by measuring routing accuracy across task categories and difficulty levels. Correct routing is critical because an incorrect agent assignment can directly lead to downstream execution failure. As shown in Table~\ref{tab:task1}, the Multi-Agent design matches or outperforms the Single-Agent baseline across all categories, achieving 100\% accuracy on SIA--Easy, SIA--Hard, and PIA--Easy, and showing higher and more stable performance on PIA--Hard (96.2\%$\pm$5.66\% vs.\ 90.4\%$\pm$15.65\%). These results show that both systems achieve strong routing performance on MistyPilot-Route100, while MistyPilot provides a clearer advantage in stability and accuracy, particularly on the more challenging PIA--Hard category. This suggests that architectural role separation can better support reliable routing than collapsing all functions into a single agent.
\begin{table}[t]
\centering
\renewcommand{\arraystretch}{1.2}
\caption{Task routing correctness on Route100.}
\label{tab:task1}
\begin{tabular*}{\columnwidth}{@{\extracolsep{\fill}}lcc@{}}
\toprule
\textbf{Condition} & \textbf{MistyPilot (Multi-Agent)} & \textbf{Single-Agent} \\
\midrule
SIA -- Easy & $100\% \pm 0.00\%$  & $99.2\% \pm 1.79\%$ \\
SIA -- Hard & $100\% \pm 0.00\%$  & $100\% \pm 0.00\%$ \\
PIA -- Easy & $100\% \pm 0.00\%$  & $100\% \pm 0.00\%$ \\
PIA -- Hard & $96.2\% \pm 5.66\%$ & $90.4\% \pm 15.65\%$ \\
\bottomrule
\end{tabular*}
\end{table}

\subsection{Evaluation of PIA Performance}
We evaluate the \textit{PIA} on the MistyPilot-SensorBind40 suite, focusing on whether it can correctly establish dynamic sensor-skill bindings and execute direct skill invocations. We follow the same comparison setting as above. As shown in Table~\ref{tab:task2}, both designs achieve 100\% accuracy on the Easy subset, indicating that both can handle straightforward physical-interaction tasks. On the Hard subset, however, MistyPilot performs substantially better and more stably, reaching 100.00\%$\pm$0\% compared with 81.00\%$\pm$4.18\% for the Single-Agent baseline. These results show that while both systems perform well on simple cases, MistyPilot provides stronger reliability for more challenging sensor-skill binding and sensor-free skill invocation tasks.

\begin{table}[t]
  \centering
  \renewcommand{\arraystretch}{1.2}
  \caption{PIA performance on SensorBind40.}
  \label{tab:task2}
  \begin{tabular*}{\columnwidth}{@{\extracolsep{\fill}}lcc@{}}
    \toprule
    \textbf{Subset} & \textbf{MistyPilot (Multi-Agent)} & \textbf{Single-Agent} \\
    \midrule
    Easy & $100.00\% \pm 0.00\%$ & $100.00\% \pm 0.00\%$ \\
    Hard & $100.00\% \pm 0.00\%$ & $81.00\% \pm 4.18\%$  \\
    \bottomrule
  \end{tabular*}
\end{table}

\begin{table}[t]
  \centering
  \renewcommand{\arraystretch}{1.2}
  \caption{SIA performance on TaskParser256.}
  \label{tab:task3}
  \begin{tabular*}{\columnwidth}{@{\extracolsep{\fill}}lcc@{}}
    \toprule
    \textbf{Subset} & \textbf{MistyPilot (Multi-Agent)} & \textbf{Single-Agent} \\
    \midrule
    Easy & $99.29\% \pm 1.60\%$ & $91.43\% \pm 13.27\%$ \\
    Hard & $96.75\% \pm 1.15\%$ & $93.50\% \pm 5.36\%$  \\
    \bottomrule
  \end{tabular*}
\end{table}

\subsection{Evaluation of SIA Performance}
For the \textit{SIA}, accurate \textit{Task State} recognition is critical for multi-turn dialogue, since decisions such as \texttt{UPDATE} and \texttt{DELETE} directly affect downstream module coordination. We evaluate \textit{Task State} and \textit{System Control} correctness on the MistyPilot-TaskParser256 suite, using the same Single-Agent baseline for comparison. As shown in Table~\ref{tab:task3}, both systems perform well, but MistyPilot achieves higher accuracy with substantially lower variance on both subsets: 99.29\%$\pm$1.60\% vs.\ 91.43\%$\pm$13.27\% on Easy, and 96.75\%$\pm$1.15\% vs.\ 93.50\%$\pm$5.36\% on Hard. These results indicate that MistyPilot provides more reliable \textit{SIA} performance in long-horizon interactions involving ambiguous references, topic shifts, and evolving instructions.

\subsection{Evaluation of Fast-Thinking Retrieval}
Fast Thinking reuses user-saved results for semantically similar requests, retrieving prior implementations instead of regenerating them from scratch to reduce redundant computation and improve response speed. We evaluate this capability on the MistyPilot-FastThinking230 suite. MistyPilot decomposes each task state into \textit{main\_task} and \textit{details}, using the \textit{main\_task} representation for Fast Path retrieval. As an ablation, we compare this structured representation with a \textit{Raw Text} baseline that retrieves directly from the original user input. We report three metrics: \textit{Top-1 Accuracy}, \textit{Rank1 Dist. Mean}, and \textit{Rank2 Dist. Mean}, where the latter two measure the average embedding-space distance between the query and its top-1 and top-2 retrieved candidates. To test whether the effect is consistent across embedding spaces, we evaluate three embedding models: two proprietary models, \texttt{text-embedding-3-large} and \texttt{text-embedding-3-small}, and one open-source model, \texttt{all-MiniLM-L6-v2}.

As shown in Table~\ref{tab:test4}, MistyPilot's disentangled representation achieves 100.00\% Top-1 accuracy across all three embedding models, while the \textit{Raw Text} baseline reaches only 67.83\%, 72.61\%, and 58.70\%, respectively. The disentangled representation also yields a larger gap between \textit{Rank1 Dist. Mean} and \textit{Rank2 Dist. Mean}, indicating better separation between the correct match and the second-best candidate. In addition, the Fast-Thinking path reduces response time from $5.088 \pm 2.571$\,s to $2.263 \pm 0.627$\,s, corresponding to a 55.5\% latency reduction.

\begin{table}[t]
  \centering
  \renewcommand{\arraystretch}{1.2}
  \caption{Fast-thinking retrieval on FastThinking230: Raw Text vs. Fast Path.}
  \label{tab:test4}
  \small
  \begin{threeparttable}
  \begin{tabular*}{\columnwidth}{@{\extracolsep{\fill}}lllll@{}}
    \toprule
    \textbf{Data} & \textbf{Model} & \textbf{Top-1} & \textbf{Rank1 Dist.} & \textbf{Rank2 Dist.} \\
    \midrule
    Raw Text  & TE3-L  & 67.83\%  & $0.365 \pm 0.070$ & $0.464 \pm 0.071$ \\
    Raw Text  & TE3-S  & 72.61\%  & $0.342 \pm 0.082$ & $0.453 \pm 0.066$ \\
    Raw Text  & MiniLM & 58.70\%  & $0.397 \pm 0.127$ & $0.570 \pm 0.066$ \\
    \midrule
    Fast Path & TE3-L  & 100.00\% & $0.003 \pm 0.015$ & $0.392 \pm 0.216$ \\
    Fast Path & TE3-S  & 100.00\% & $0.003 \pm 0.015$ & $0.394 \pm 0.211$ \\
    Fast Path & MiniLM & 100.00\% & $0.000 \pm 0.002$ & $0.463 \pm 0.248$ \\
    \bottomrule
  \end{tabular*}
  \begin{tablenotes}[flushleft]
    \footnotesize
    \item \textit{Note:} TE3-L = text-embedding-3-large; TE3-S = text-embedding-3-small; MiniLM = all-MiniLM-L6-v2.
  \end{tablenotes}
  \end{threeparttable}
\end{table}

\subsection{Evaluation of Skill Extensibility}
A key goal of MistyPilot is plug-and-play skill extension: new skills can be added to the library using only their docstring descriptions. We evaluate whether MistyPilot can still discover newly added skills and select the correct one as the library scales. Specifically, we test libraries with 30, 50, 70, and 100 skills on the SkillExtension suite, repeating each configuration five times. As shown in Table~\ref{tab:test6}, MistyPilot selects the correct skill in all runs across all library sizes, indicating that docstring-based skill selection remains reliable for libraries of up to 100 skills. We observe no degradation within this range, while characterizing the upper scalability limit is left to future work.

\begin{table}[t]
  \centering
  \renewcommand{\arraystretch}{1.2}
  \caption{Skill extensibility on SkillExtension.}
  \label{tab:test6}
  \begin{tabular*}{\columnwidth}{@{\extracolsep{\fill}}lcccc@{}}
    \toprule
    \textbf{Scale} & \textbf{30 skills} & \textbf{50 skills} & \textbf{70 skills} & \textbf{100 skills} \\
    \midrule
    Correct skill selection & $\checkmark$ & $\checkmark$ & $\checkmark$ & $\checkmark$ \\
    \bottomrule
  \end{tabular*}
\end{table}

\begin{table}[h]
\centering
\renewcommand{\arraystretch}{1.3}
\caption{Subjective evaluation results of the MistyPilot prototype.}
\label{tab:user_study}
\begin{threeparttable}
\begin{tabularx}{\columnwidth}{@{}>{\raggedright\arraybackslash}X !{\color{gray}\vrule} >{\centering\arraybackslash}p{3em} !{\color{gray}\vrule} >{\centering\arraybackslash}p{3em}@{}}
\toprule
\textbf{Evaluation Dimension and Questionnaire Item} & \textbf{Mean} & \textbf{Std} \\
\midrule
\textbf{Interaction Naturalness:} Communicating with the robot through natural language to perform specific tasks was intuitive and straightforward. & 4.75 & 0.45 \\
\midrule
\textbf{Comprehension Accuracy:} The robot accurately understood my intentions and provided responses that met my expectations. & 4.42 & 0.67 \\
\midrule
\textbf{Emotional Expressiveness:} The robot's speech and reactions were emotionally expressive, giving it a highly anthropomorphic and human-like presence. & 4.75 & 0.45 \\
\midrule
\textbf{System Responsiveness:} The overall response latency was within an acceptable range. & 4.67 & 0.49 \\
\midrule
\textbf{User Satisfaction:} Overall, I am satisfied with the multimodal interaction experience provided by the MistyPilot system. & 4.75 & 0.45 \\
\bottomrule
\end{tabularx}
\begin{tablenotes}[flushleft]
\footnotesize
\item \textit{Note:} All scores are on a 5-point Likert scale (1 = strongly disagree, 5 = strongly agree). Mean is the average rating; Std is the standard deviation.
\end{tablenotes}
\end{threeparttable}
\end{table}

\subsection{Preliminary User Feedback Study}
As a preliminary assessment of practical feasibility and user perception, we conducted an in-person user study with 12 volunteers. Each participant interacted with MistyPilot for at least 30 minutes, totaling over 6 hours of interaction. Participants used SIA for open-domain conversations and PIA for 30 utility skills. Afterward, they completed a five-item post-study questionnaire (Table~\ref{tab:user_study}). The study was approved by our institution's IRB.

The questionnaire was adapted from established HRI and UX instruments and tailored to our setting ~\cite{bartneck2009measurement}. Participants rated five aspects on a 5-point Likert scale.
Results are consistently positive, with all mean scores above 4.4/5. Overall satisfaction, interaction naturalness, and emotional expressiveness received the highest ratings (Table~\ref{tab:user_study}). Occasional failures were mainly caused by ASR errors on accented speech, leading to misunderstood user intent.

\section{Limitations and Future Work}
Our evaluation suites target the components we consider most critical, but they are not exhaustive: they do not cover every interaction pattern a social robot may encounter, and designing broader evaluation suites, including composite and longer-horizon tasks, remains future work. In addition, sensor-skill bindings and user-marked results currently persist across sessions. While this supports continuity for individual users, it may raise privacy concerns on shared robots. Future work will introduce user authentication together with per-user isolation of persistent bindings and long-term memory.

\section{Conclusion}

We presented MistyPilot, a multi-agent LLM framework for natural-language skill orchestration on the Misty social robot. Its role-specialized architecture separates reactive physical interaction from stateful social interaction through the \textit{PIA} and \textit{SIA}. MistyPilot supports sensor-triggered skill execution, dialogue-oriented task-state management, result reuse, multimodal response generation, and extensibility to new skills -- capabilities that existing interfaces expose only through hand-written code.
Evaluations demonstrate higher component-level accuracy and lower variance than the single-agent baseline, while a preliminary user study indicates positive perceptions of usability and interaction quality.

\section{Acknowledgment}
This material is based upon work supported under the AI Research Institutes program by the U.S. National Science Foundation and the Institute of Education Sciences, U.S. Department of Education, through Award \# 2229873—National AI Institute for Exceptional Education. Any opinions, findings and conclusions, or recommendations expressed in this material are those of the author(s) and do not necessarily reflect the views of the National Science Foundation, the Institute of Education Sciences, or the U.S. Department of Education.

\newpage

\par
\vspace{0.6\baselineskip} 

\bibliographystyle{splncs04}
\bibliography{main}
\end{document}